%% file: root.tex
\documentclass[letterpaper, 10 pt, conference]{ieeeconf}  

\IEEEoverridecommandlockouts                              

\usepackage{microtype}
\usepackage{graphicx} 
\usepackage{subcaption} 
\usepackage{todonotes}
\usepackage{amsmath} 
\usepackage{color}
\usepackage{colortbl}
\usepackage{soul} 
\usepackage{xcolor}
\usepackage{comment}
\usepackage{multicol}
\usepackage{multirow}
\usepackage{hyperref}

\usepackage[separate-uncertainty=true,multi-part-units=single]{siunitx} 
\DeclareSIUnit\decade{dec}

\newcommand{\rev}[1]{\textcolor{black}{#1}}
\newcommand{\revM}[1]{\textcolor{black}{#1}}

\usepackage[pages=some,placement=top, scale=1,
hshift=0,
vshift=-20,]{background}
\usepackage{lipsum}
\backgroundsetup{contents=This paper has been accepted for publication at 2025 IEEE International Conference on Advanced Robotics (ICAR),firstpage=true, angle=0,color=gray, opacity=1}

\title{\LARGE \bf
Tactile Gesture Recognition with Built-in Joint Sensors for Industrial Robots
}

\author{Deqing Song$^{1*}$, Weimin Yang$^{1*}$, Maryam Rezayati$^{1, 2\dagger}$, Hans Wernher van de Venn$^{2\dagger}$ 
\thanks{Authors indicated by $^1$ are with Dep. of Informatics, University of Zurich Switzerland. {\tt\small FirstName.LastName@uzh.ch}}
\thanks{Authors indicated by $^2$ are with the Institute of Mechatronics systems, Zurich University of Applied Science, Switzerland. {\tt\small FirstName.LastName@zhaw.ch}}
\thanks{Authors marked with * contributed equally and marked with $\dagger$ are corresponding authors. This work was supported by the Eurostars project (Grant No. E!3087) titled \textit{SmartSenseAI}. The authors would like to thank generative AI models (ChatGPT and Gemini) for assisting with improving the clarity of the manuscript.}
\thanks{The code and dataset corresponding to this work is provided at \href{https://github.com/MindLabZHAW/tactileGestureDetection}{https://github.com/MindLabZHAW/tactileGestureDetection}}
}

\begin{document}

\maketitle

\begin{abstract}
While gesture recognition using vision or robot skins is an active research area in Human-Robot Collaboration (HRC), this paper explores deep learning methods relying solely on a robot’s built-in joint sensors, eliminating the need for external sensors. We evaluated various convolutional neural network (CNN) architectures and collected a dataset to study the impact of data representation and model architecture on the recognition accuracy. Our results show that spectrogram-based representations significantly improve accuracy, while model architecture plays a smaller role. We also tested generalization to new robot poses, where spectrogram-based models performed better. Implemented on a Franka Emika Research robot, two of our methods, STFT2DCNN and STT3DCNN, achieved over 95\% accuracy in contact detection and gesture classification. These findings demonstrate the feasibility of external-sensor-free tactile recognition and promote further research toward cost-effective, scalable solutions for HRC.
\end{abstract}

\begin{figure}
    \centering
    \includegraphics[width=1\linewidth]{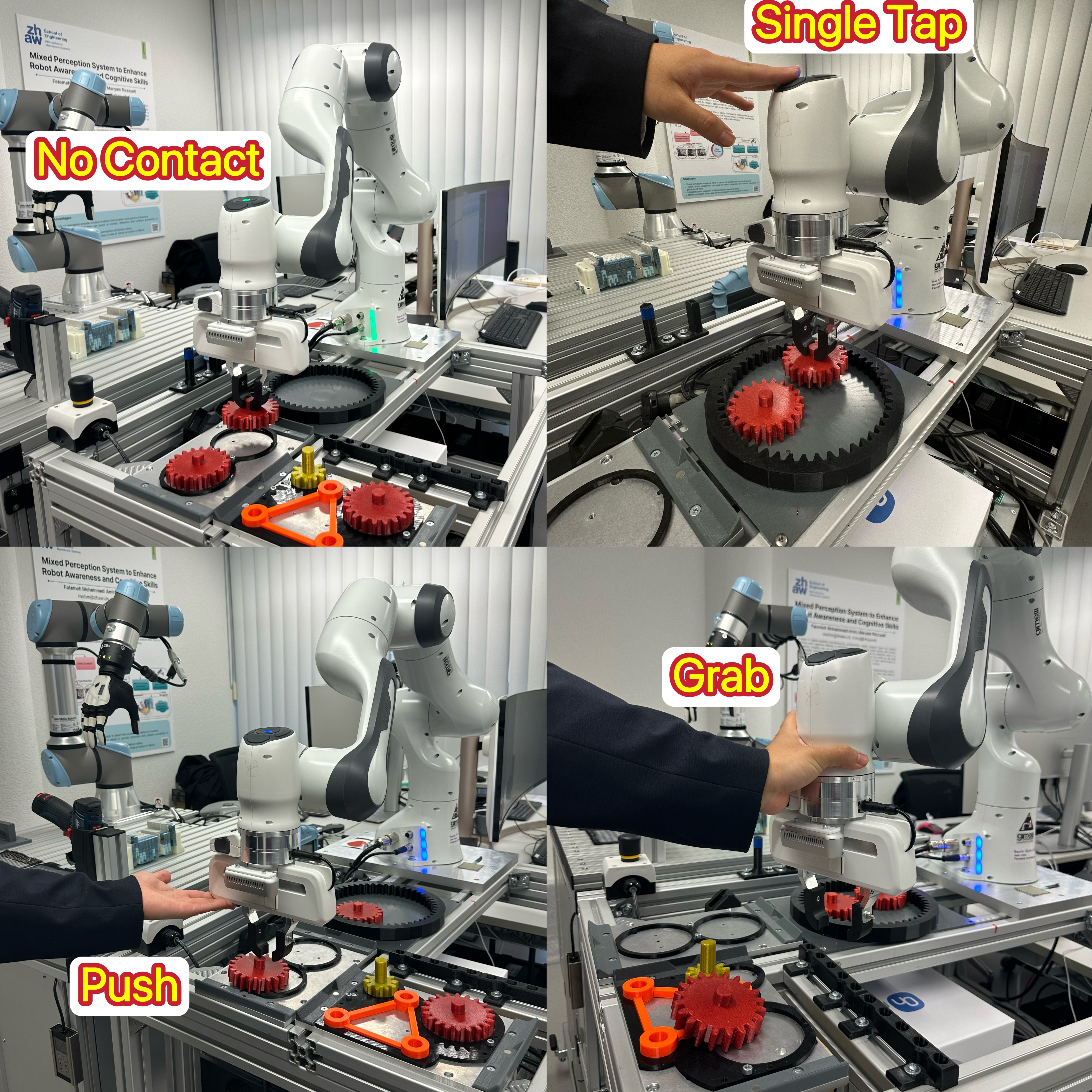}
    \caption{Application scenario of the proposed tactile gesture recognition system during a gear assembling task.}
    \label{fig:headingimage}
\end{figure}

\section{INTRODUCTION}\label{sec:introduction}

\revM{Industry is shifting from Industry 4.0 to Industry 5.0, emphasizing worker well-being in production \cite{Xu2021, Maddikunta2021, Leng2022}. When humans control tasks, they feel more comfortable if they can influence robot movements \cite{rubagotti2022perceived}, prompting research on communication in human-robot collaboration, especially through gestures and touch \cite{pohlt2018effects}.}
\revM{Tactile perception in robotics often uses artificial skins with sensors detecting touch and pressure. These sensors fall into two categories: continuous analog sensors \cite{Barreiros2022} and neuromorphic spike-based devices mimicking biological skin \cite{Hu2023}. Analog sensors face challenges with high data volume as sensor density increases \cite{Liu2022}, leading to interest in spike-based signals \cite{Wang2023}. However, both types pose design and integration challenges due to complex hardware and the need for external systems \cite{Shih2020, Huang2024, Heo2019, Lee2016, Zaatari2019, Zech2023}, limiting their deployment in large-scale manufacturing.}
\revM{To address these issues, recent work explores using robots’ built-in sensors, such as joint torque sensors, for tactile recognition. These sensors provide real-time feedback on robot-environment interaction without added hardware and complexity. However, most joint-torque-based studies focus on collision detection rather than recognizing multiple human touch gestures.}
\hypertarget{reviewer1_2}{\rev{In this context, tactile gesture recognition can be viewed as a multi-channel temporal classification problem, where time-varying joint-level state signals across multiple joints must be recognized in real time—posing significant challenges for conventional rule-based approaches.}}


\subsection{The Paper's Contribution}
Concretely, we make the following contributions: 
\begin{itemize}
    \item We demonstrate the feasibility of recognizing multiple tactile gestures on a robot solely by using internal joint sensor data, eliminating the need for expensive and complex external tactile sensors.
    \item We introduce a tactile gesture recognition framework that combines spectrogram-based transformations of internal joint sensor data with CNN-based classification, effectively bridging the gap between time-domain signal processing and image-based deep learning.
    \item We conduct a comprehensive exploration of signal representations and implement a variety of strategies to capture the distinct signal patterns generated by tactile gestures.
    \item We compare and evaluate the performance of these methods in real-time on a Franka Emika Research robotic platform, validating their practical applicability.
    \item To ensure the reproducibility and facilitate further research, we have made our complete dataset and source code publicly available.
\end{itemize}

\section{Related work}\label{sec:RelatedWork} 

\subsection{Collision Detection}
\revM{Lee et al. \cite{Lee2016} combined dynamic and friction models to monitor accelerations, while Ren et al. \cite{Ren2018} used an extended state observer to estimate external torque. Zurlo et al. \cite{Zurlo2023} tracked energy and momentum before applying a Contact Particle Filter. These model-based methods require extensive modeling, threshold tuning, and are sensitive to noise, motivating model-free approaches.}

\revM{Sharkawy et al. \cite{Sharkawy2019} replaced explicit dynamic modeling with multi-layer neural networks trained using the Levenberg–Marquardt algorithm. In later work \cite{Sharkawy2022}, they compared the mentioned model with cascaded forward neural networks (CFNN) and RNNs, finding the single-hidden-layer neural network most accurate but dependent on torque sensors, while CFNN and RNN performed well without them. To eliminate manual threshold tuning, Park et al. \cite{KMPark2021, KMPark2022} integrated Momentum Observers with CNNs and SVMs for supervised and unsupervised detection. Min et al. \cite{Min2019} employed MLP based on vibration data from joint sensors, and Heo et al. Leveraging the full suite of modern cobots' joint-sensor data (such as external torques and accelerations), \cite{Heo2019} proposed CollisionNet, an AR-based CNN for direct collision prediction. Fathi et al. \cite{Fathi2022GPR} combined a Gaussian Process Regressor (GPR) for external torque estimation with a CNN for contact classification, requiring retraining only of the GPR to maintain accuracy across different motions and speeds. Rezayati et al. \cite{rezayati2022} used deep metric learning to enhance model generalization across similar robots, while Czubenko et al. \cite{Czubenko2021} and Kim et al. \cite{DKim2022} developed ConvLSTM and a versatile modularized neural network (MNN) for improved dynamic modeling and transferability.}
\subsection{Extended classification and tactile recognition}
Beyond collision detection, researchers have also examined collision classification \cite{Min2019}. For example, Min et al. \cite{Min2019} distinguished between no-contact scenarios and collisions at two points in three directions, while Amin et al. \cite{Amin2020} used a 1D-CNN to classify no-contact conditions alongside intentional and incidental contact on two distinct links.

Similarly, in the realm of tactile gesture recognition, which deals with active contact events, the cobot must first detect the contact, then identify which gesture is applied. Bianchini et al. \cite{Bianchini2021} defined four types of gestures — Tap (e.g., hit, pat, poke, slap), Touch (e.g., contact, lift, press), Grab (e.g., cradle, pinch, squeeze), and Slip (e.g., rub, scratch, stroke) — and evaluated several classification approaches, including traditional machine learning methods (Decision Trees, SVMs), and a 1D CNN network. Their experiments, conducted on a UR5e cobot equipped with the internal six-axis wrist force-torque sensor, demonstrated that the neural network classifier achieved the best performance, with a test accuracy of \hypertarget{reviewer1_1}{\rev{\textbf{81\%}}}. Our work builds upon this foundation by investigating a specific deep learning approach. We further investigate CNN-based deep learning methods on various 3D data representations (including spectrograms), and provide a detailed analysis of how input representations and network architectures influence performance, particularly under domain shifts across unseen robot poses, a crucial challenge for real-world generalization.%



\section{Material And Methods}\label{sec:Methods}

\input{tabels/1_NWStructures}
\begin{figure*}[htb]
    \begin{center}
        \includegraphics[width=1\textwidth]{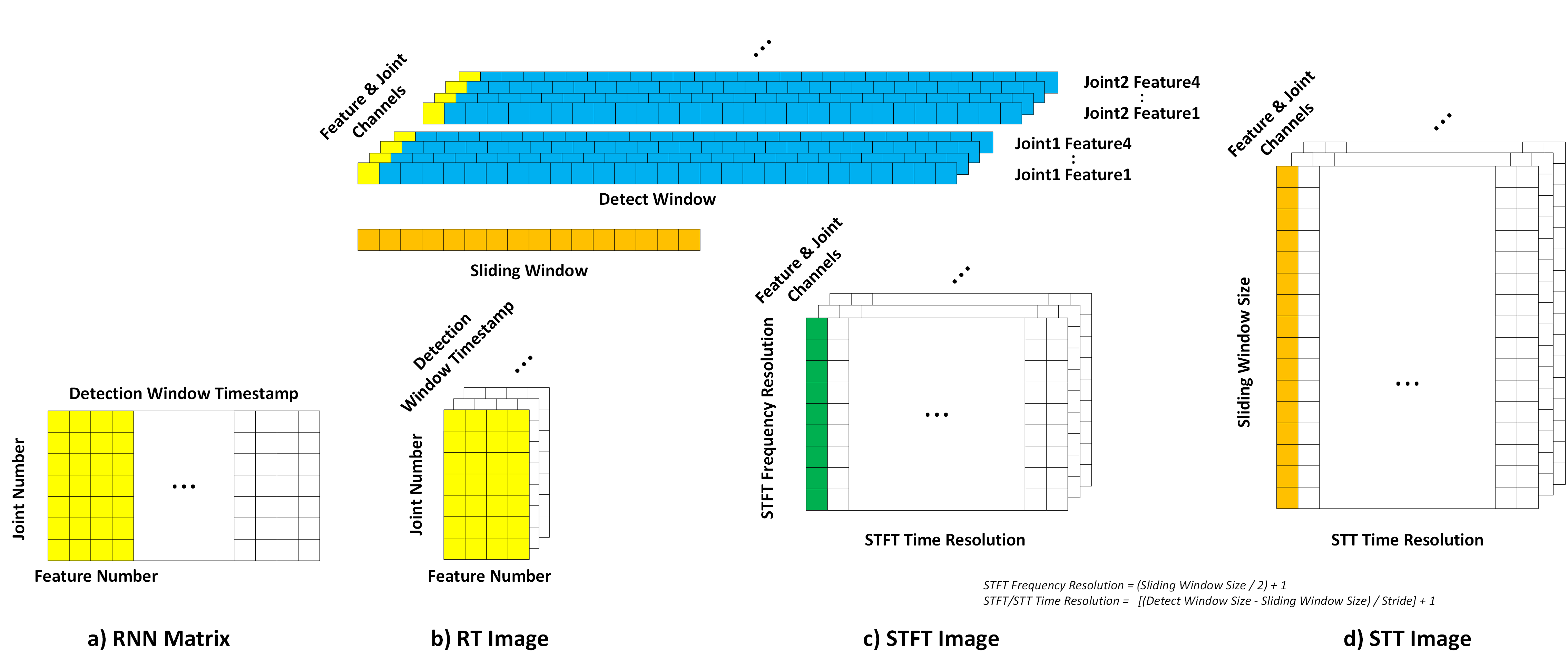}
        \caption{The construction of all input formats of our methods. Illustrate as parameter settings: Joint Number = 7, Feature Number = 4 ($e, \dot e, \tau, \tau_{ext}$), Detect Window Size = 28, Sliding Window Size = 16}
        \label{fig:input}
    \end{center}
\end{figure*}
\subsection{Problem Statement}
Tactile gesture recognition can be formulated as a multi-channel signal classification problem. When a human operator interacts with the robot using different gestures, the resulting sensor signals vary uniquely across joints, capturing distinct temporal and spatial patterns. These signals are typically high-dimensional, non-linear, and gesture-dependent in duration, making real-time classification challenging. Traditional rule-based or threshold-based methods struggle to handle such variability, necessitating the need for machine learning or deep learning approaches to effectively recognize and classify tactile gestures.
This paper addresses three key research questions:
\begin{itemize}
    \item How can a CNN-based algorithm be \textbf{designed} to recognize time-varying tactile gestures from internal robot sensors?
    \item What is the impact of different \textbf{input representations} (e.g., spectrograms vs.\ raw signals) on performance?
    \item How do the proposed models \textbf{perform}, and what factors explain performance differences?
\end{itemize}

\subsection{Dataset Collection}\label{subsec:dataset}
Prior work has categorized human-to-robot touch gestures into different taxonomies, ranging from 30 \cite{Yohanan2012} to as few as 4 \cite{Bianchini2021}. \hypertarget{reviewer1_4}{\rev{In line with these studies, especially \cite{Bianchini2021}, we define three key gestures (Single Tap, Push, Grab) that sufficiently capture essential touch interactions for our demonstrator. These gestures were chosen because they span a range of contact dynamics, from brief impulsive touches to sustained directional forces and stable multi-directional forces.}}
\textbf{\textit{Single Tap}} is a brief, impulse-like touch applied to the robot surface without sustained pressure, such as pat, poke, or slap. 
\textbf{\textit{Push}} is a sustained force applied to the robot with a clear direction, such as push, pull, or lift. 
\textbf{\textit{Grab}} is a prolonged contact in which the hand encloses a link of the robot and applies stable pressure from multiple directions, such as pinch or squeeze. 

Data were captured at a sampling frequency of 200~Hz using a digital glove and interacting with robot hand and the connected end effector in 5 directions (left, right, front, back, and up) and 10 positions\hypertarget{reviewer1_3}{\rev{, to ensure balanced data coverage and avoid directional bias.}} Data are segmented into fixed detection time windows of 28 timestamps with a step size of 14\hypertarget{reviewer1_3}{\rev{, balancing real-time responsiveness, recognition accuracy, and data redundancy.}}

\subsection{Proposed Gesture Recognition Approach}

This section presents the proposed models for real-time tactile gesture recognition on a Franka Emika robot, using only internal joint sensors. Various data representations are explored as model inputs.

\subsubsection{Traditional Machine Learning Methods}
\hypertarget{reviewer_1_5}{\rev{As baselines, we include a traditional K-Nearest Neighbor (KNN) classifier and a Long Short-Term Memory (LSTM) network. KNN serves as a lightweight, non-parametric reference, while LSTM provides a recurrent baseline capturing temporal dependencies in the data. Both methods often achieve competitive results with lower computational cost and reduced overfitting risk compared to more complex architectures.}}

\subsubsection{CNN-Based Deep Learning Models}
Convolutional neural networks (CNNs) are effective in capturing temporal patterns in physical interaction data \cite{Heo2019, KMPark2021, Czubenko2021, Amin2020}. Table~\ref{tab:nwstructures} summarizes the proposed architectures, \hypertarget{reviewer1_6}{\rev{which adopt a conventional pipeline with an input layer, convolutional and pooling layers, a flatten layer, and fully connected layers.}}

\subsubsection{Input Data Representation}

For our gesture classification tasks, we leverage time-frequency analysis to handle the non-stationary nature of the sensor signals \cite{Yang2019}. At each time step $t$, the feature vector for joint, $i$ is defined as
\[
\mathbf{j}^i_t = \left[ e^i_t, \ \dot{e}^i_t, \ \tau^i_t, \ \tau^{i,\mathrm{ext}}_t \right],
\]
where $e^i_t$ is the joint position error, $\dot{e}^i_t$ is the joint velocity error, $\tau^i_t$ is the measured joint torque, and $\tau^{i,\mathrm{ext}}_t$ is the estimated external torque. We explore three distinct 3D data representations to serve as inputs for our models, each designed to test the significance of key signal properties:
\begin{itemize}
\item Whether frequency-domain features are necessary for accurate classification.
\item Whether a pre-processing sliding-window operation is beneficial.
\end{itemize}

\textbf{Non-Spectrogram (Raw-Time Stack):}
To serve as a baseline and investigate the impact of the sliding-window pre-processing itself, we construct a Raw-Time Stack. This representation is a 3D temporal structure formed by simply stacking the full time-series of joint-by-feature matrices, without any pre-processing windowing or transformation (Figure~\ref{fig:input}(b)).

\textbf{Spectrogram:}
We generate spectrograms using the Short-Time Fourier Transform (STFT), a common time-frequency analysis technique for non-stationary signals \cite{Yang2019}. We chose STFT over the more computationally demanding continuous wavelet transform to meet the real-time requirements of our system. To generate our spectrograms, we use a sliding-window operation on each feature and then apply the STFT. The resulting 2D spectrograms are stacked into a 3D structure with dimensions for frequency, time, and combined feature-joint indices, as illustrated in Figure~\ref{fig:input}(c).

\textbf{Pseudo-Spectrogram:}
To specifically investigate whether frequency-based features are critical for performance, we introduce a Pseudo-Spectrogram (Figure~\ref{fig:input}(d)). This representation is constructed using the same sliding-window process as the STFT, thereby preserving the 3D input shape. However, it replaces the frequency-domain data with raw time-domain values.  Comparing this representation with the Spectrogram will directly reveal the importance of frequency information.

\subsubsection{Majority Voting}
As noted by Czubenko et al. \cite{Czubenko2021}, filtering multiple consecutive detections can effectively reduce false positives. Therefore, for real-time deployment, we apply a majority-voting scheme over three consecutive detection windows for all our models, helping mitigate jitter in the classification results.

\subsection{Hardware and Software Setup}

\begin{figure*}
    \begin{center}
        \includegraphics[width=1\textwidth]{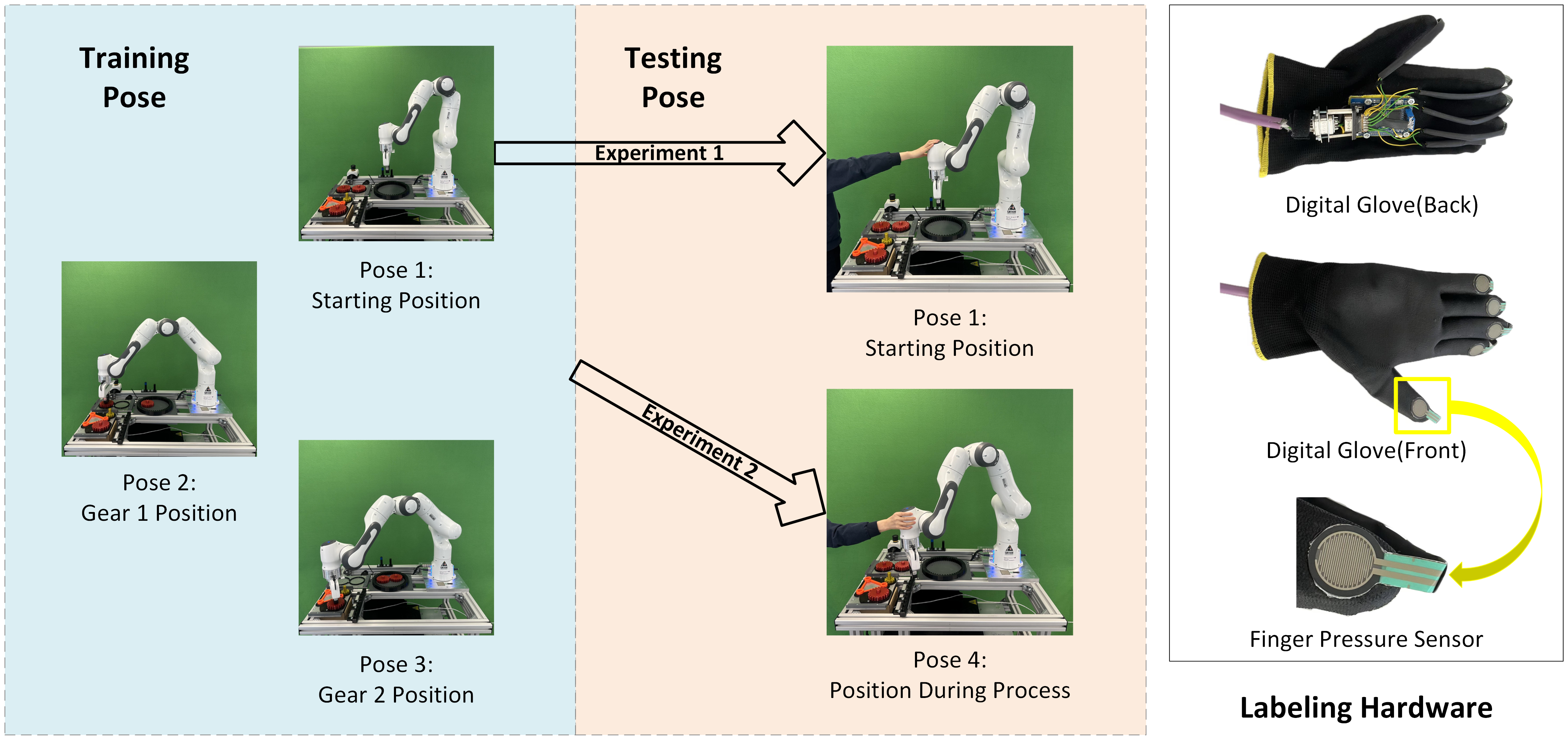}
        \caption{Experimental Setup (Gear Assembling Task): Left part shows 3 poses in the assembling procedure that were used to collect training data; Middle part shows 2 test poses used in experiments 1 and 2; Right part shows the digital glove and its pressure sensor}
        \label{fig:exsetup}
    \end{center}
\end{figure*}

The experimental platform consists of a Franka Emika Research collaborative robot, a digital glove with pressure sensors (Fig. \ref{fig:exsetup}), and a computer running Ubuntu 20.04 LTS with a real-time kernel. Pressure sensors on the glove detect contacts between the robot and a human, sending signals to the computer via the CANBUS protocol with a latency of less than 600~$\mu$s. The computer has 32~GB RAM and Intel Core i7 CPU.

\section{Experimental Results}\label{sec:Results}

\subsection{Evaluation Metrics}\label{subsec:metrics}

\begin{figure}
    \centering
    \includegraphics[width=1\linewidth]{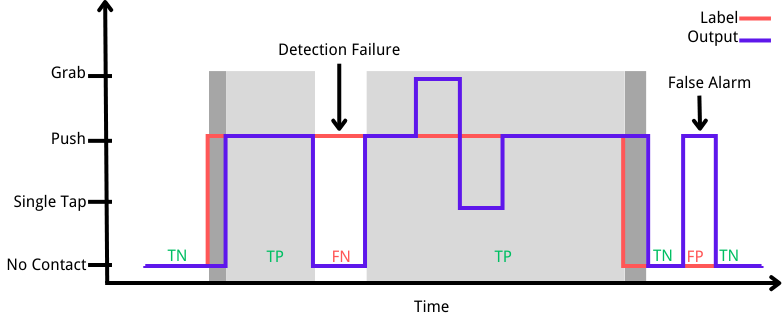}
    \caption{Evaluation Example: A Single Push. True Negative (TN): The model correctly identifies no contact when there is no contact. True Positive (TP): The model correctly detects a contact event when contact occurs. False Negative (FN): The model fails to detect contact when contact actually occurs. False Positive (FP): The model falsely detects contact when no contact occurs.}
    \label{fig:contact_metrics}
\end{figure}

To evaluate model performance, we follow the methodology in \cite{Rezayati2025LSTM}, assessing two aspects: (1) the ability to accurately detect contact events, and (2) the ability to correctly classify tactile gesture types.

\subsubsection{Contact Detection Metrics}
Since reliable contact detection is the basis for tactile gesture recognition, we quantify performance using the following metrics:
\begin{itemize}
    \item \textbf{Accuracy (ACC, \%)}: Overall correctness in detecting contact events:
    \begin{equation}
        \mathrm{ACC} = \frac{\mathrm{TP} + \mathrm{TN}}{\mathrm{TP} + \mathrm{TN} + \mathrm{FP} + \mathrm{FN}}.
    \end{equation}
    
    \item \textbf{Detection Failure Rate (DFR, \%)}: Proportion of actual contact events missed:
    \begin{equation}
        \mathrm{DFR} = \frac{\mathrm{FN}}{\mathrm{TP} + \mathrm{FN}}.
        \label{eq:detection_failure_rate}
    \end{equation}

    \item \textbf{False Alarm Rate (FAR, \%)}: Proportion of incorrect contact detections:
    \begin{equation}
        \mathrm{FAR} = \frac{\mathrm{FP}}{\mathrm{TN} + \mathrm{FP}}.
    \end{equation}
    
    \item \textbf{Detection Delay (DD, ms)}: The time required for the model to recognize a contact occurrence.
    
    \item \textbf{Recovery Delay (RD, ms)}: The time required for the model to recognize the end of a contact event.
\end{itemize}
As noted in \cite{Rezayati2025LSTM}, lower DFR and FAR indicate a more robust detector. Delays exceeding 150~ms in either DD or RD are treated as misclassifications.

\subsubsection{Gesture Classification Metrics}

Gesture classification performance is assessed only within correctly detected contact regions (True Positive segments in Fig.~\ref{fig:contact_metrics}) to avoid bias from detection errors.
We report the following metrics:

\begin{itemize} 
\item \textbf{Accuracy (ACC, \%):} Measures the overall correctness in classifying gesture types, defined as: 
\begin{equation} 
\text{ACC} = \frac{\text{TP}_{\text{gesture}}}{{\text{TP}}_{\text{contact}}}. 
\end{equation}

\item \textbf{Per-gesture Detection Failure Rate (\%):} Contact localization failure rate for individual gesture defined as Eq \ref{eq:detection_failure_rate}.

\end{itemize}



\subsection{EXP1: Gesture Detection in a Known Robot Pose}
\input{tabels/2_Ex1Result}
This experiment evaluates the models’ ability to classify gesture types in a fixed robot pose (Pose~1, Fig.~\ref{fig:exsetup}). Training and testing are conducted on the same pose to assess accuracy in distinguishing gestures. 
The KNN baseline was excluded after preliminary testing, as it failed to reliably separate gestures from one another or from no-contact cases. In addition, increasing training data caused cumulative delays that violated real-time constraints. 
Final results are presented in Table~\ref{tab:exp1_results}. STFT2DCNN and STT3DCNN models achieve the highest gesture classification accuracy while maintaining competitive contact detection performance.

\subsection{EXP2: Cross-Pose Generalization}
This experiment evaluates generalization to unseen robot poses (Fig.~\ref{fig:exsetup}). Models are trained on data from Poses~1–3 and tested on Pose~4, introducing a distribution shift.
Results in Table~\ref{tab:exp1_results} show that spectrogram-based models generalize significantly better than non-spectrogram approaches. Compared with EXP1, the reduction in contact detection and gesture classification accuracy is minimal—and in some cases, accuracy improves. This indicates that frequency-domain features captured by spectrograms are more robust to pose variation. This shows that features captured by constructing spectrogram or pseudo-spectrogram representations yield models that are more robust to pose variation. For example, STFT3DCNN’s contact detection accuracy increases from 94.42\% to 96.22\%, while its gesture classification accuracy rises slightly from 93.05\% to 93.08\%.

\subsection{EXP3: Impact of 3D Convolutions and Frequency Transformations}
This experiment investigates whether 3D convolutions and frequency-domain transformations (e.g., STFT) influence performance. We compare STFT2DCNN and STT2DCNN (\textbf{2D CNNs}) with STFT3DCNN and STT3DCNN (\textbf{3D CNNs}) using identical network architectures.

\input{tabels/4_EXP3Comparison}

The STFT3DCNN and STT3DCNN results in Tables~\ref{tab:exp1_results} correspond to \textbf{STFT3DCNN7+1} and \textbf{STT3DCNN4+7}, respectively.
From Table~\ref{tab:exp3_comparison}, we find that \textbf{under identical architecture}, introducing \textbf{3D convolution} can reduce gesture classification accuracy (e.g., STFT/STT2DCNN vs.\ STFT/STT3DCNN7+1). However, slightly deepening the 3D CNN (STT3DCNN7+1 $\rightarrow$ STT3DCNN4+7) improves performance, even surpassing STT2DCNN in Experiment~2.

Increasing STFT3DCNN complexity (STFT3DCNN7+1 $\rightarrow$ STFT3DCNN4+7) introduces substantial cumulative delay, which prevented evaluation of accuracy and is therefore indicated as “–” in the table. This delay disappears on a high-performance GPU workstation, indicating that real-time viability depends strongly on available hardware.
In conclusion, both frequency transformations (STT $\rightarrow$ STFT) and 3D convolutions incur extra computation and have a relatively substantial impact on model performance.
\section{Discussion}\label{sec:Discussion}

\subsection{Result Analysis and Answering Research Questions}
\textbf{RQ1:} 
Our experiments validate the design of a CNN-based framework for tactile gesture recognition \hypertarget{reviewer1_8}{\rev{as all CNN models achieve competitive accuracy (above 90\%)}}. The high performance of our leading models, such as STFT2DCNN and STT3DCNN, with over 95\% accuracy in contact detection and gesture classification, demonstrates the effectiveness of our design.

\textbf{RQ2:}
Our results clearly demonstrate that spectrogram-based methods significantly outperform non-spectrogram methods in tactile gesture recognition. This indicates that constructing spectrogram or pseudo-spectrogram representations — through sliding-window segmentation and column-wise stacking — is highly beneficial when combined with CNN architectures. However, for simpler tasks like only contact detection, less complex non-spectrogram methods are sufficient.

\textbf{RQ3:}
To understand the factors contributing to these results, we compared CNNs with different architectures. Our findings show that applying 3D convolutions can sometimes decrease gesture classification performance under identical network structures. However, a slightly deeper 3D CNN architecture can improve accuracy, sometimes surpassing the 2D CNNs. Both frequency transformations (like STFT) and 3D convolutions introduce additional computational costs. This highlights a key trade-off between model complexity, task demands, and available computational resources.
Therefore, \textbf{the decision to use them should be task-dependent}:

\begin{itemize}
    \item For complex gesture recognition tasks requiring higher accuracy, well-chosen spectrogram-based representations and carefully designed 3D CNNs may be justified.
    \item For simpler tasks or systems with limited computational capacity, \textbf{2D CNNs combined with STFT}, or \textbf{3D CNNs without frequency transformations}, already offer an effective trade-off between accuracy and efficiency.
\end{itemize}

In summary, \textbf{system designers should carefully balance model complexity, task demands, and available computational resources} when designing tactile gesture recognition systems.




\subsection{Limitations and Outlook}
Our experiments were conducted under specific robot poses (Pose 1, 2, 3 for training and Pose 4 for testing), designed based on our gear-assembling task. This constrained experimental setup limits the pose generalization ability of the proposed models. In real-world applications, robots are expected to handle tactile gestures under a wider variety of poses. Although we evaluated cross-pose generalization to some extent (by testing on unseen Pose 4), the performance under poses that significantly deviate from the training trajectory remains unknown and could degrade.
Moreover, contact location generalization is also a concern. All gestures were applied solely to link 7 and the end-effector as required by the task. Consequently, our models may have implicitly overfitted to these specific robot parts, potentially leading to poor performance if tactile gestures are applied to other parts of the robot.
\hypertarget{reviewer2_1}{\rev{Additionally, all experiments were conducted on a single robotic platform — the Franka Emika Research. While representative, the model's transferability to other cobots, such as the UR series, has not been evaluated. Since hardware characteristics (e.g., joint sensor precision, dynamic response) vary, this is a crucial limitation for broader industrial deployment.}}

Although the total amount of training data is considerable, our dataset still suffers from imbalances and coverage limitations. In Experiment 1 (Ex1), which used only Pose 1 for training, the number of complete gesture instances (ST, P, G) was limited across the 10 positions. As a result, some models may exhibit "blind spots" in recognizing gestures performed at underrepresented positions.
This issue was partially alleviated in Experiment 2 (Ex2), where combining training data from three poses (Pose 1, 2, and 3) increased the diversity of gesture samples. This improvement correlates with the significant performance gains observed in models like STT2DCNN, whose accuracy notably improved from Ex1 to Ex2. This highlights the importance of gesture variability and pose diversity in training data for robust tactile recognition.

\hypertarget{reviewer2_4}{\rev{Our proposed methods focus exclusively on CNN-based architectures, without exploring more recent alternatives such as Transformer-based models. While Transformers have shown strong performance in various sequence modeling tasks, their application to tactile signals could offer advantages in capturing complex spatiotemporal patterns. However, such models typically require very large datasets. Exploring Transformer-based or hybrid architectures is a promising direction for future work that could lead to improved accuracy, better generalization, or swifter response.}}

\hypertarget{reviewer1_9, reviewer2_1}{\rev{Future work will also address practical deployment in Industry 5.0, considering latency, training efficiency, and energy use, as well as safety, robustness, and reliability under varying payloads and real-world conditions.}}




\bibliographystyle{IEEEtran}
\bibliography{ref}


\end{document}

%% file: tabels/1_NWStructures.tex
\begin{table*}[htb]
    \centering
    \renewcommand{\arraystretch}{1.2} 
    \caption{Proposed CNN-based models with their inputs, layer configurations, and key hyperparameters. \textit{(For Convolution Layer we only mentioned kernel size as paddings are all 0 and strides are all 1)}}
    \label{tab:nwstructures}
    
    \begin{tabular}{|c|ccc|}
        \hline
        \multirow{2}{*}{\centering \textbf{Model Name}} 
        & \multicolumn{3}{c|}{\textbf{Model Details}} \\
        \cline{2-4}
         & \textbf{Input} & \textbf{Layer Num.} & \textbf{Key Hyperparameters} \\
        \hline

         STFT2DCNN      & 3D STFT Image & 5 & 3DConv1($28 \times 3 \times 3$) + 3DConv2($1 \times 3\times 3$) + 3D Pool($1\times1\times1$) + Flatten + FC \\
        STFT3DCNN    & 3D STFT Image & 5 & 3DConv1($7 \times 3 \times 3$) + 3DConv2($1 \times 3 \times 3$) + 3D Pool($1 \times 1 \times 1$) + Flatten + FC \\
         STT2DCNN   & 3D STT Image  & 5 & 3DConv1($28 \times 3 \times 3$) + 3DConv2($1 \times 3 \times 3$) + 3D Pool($1 \times 1 \times 1$) + Flatten + FC \\
         STT3DCNN   & 3D STT Image  & 5 & 3DConv1($4 \times 3 \times 3$) + 3DConv2($7 \times 3 \times 3$) + 3D Pool($1 \times 1 \times 1$) + Flatten + FC \\
         RT2DCNN   & 3D RT Image   & 4 & 3DConv($28 \times 3 \times 3$) + 3D Pool($1 \times 1 \times 1$) + Flatten + FC \\
         RT3DCNN   & 3D RT Image   & 5 & 3DConv1($5 \times 3 \times 3$) + 3DConv2($5 \times 3 \times 3$) + 3D Pool($1 \times 1 \times 1$) + Flatten + FC \\
        \hline
    \end{tabular}
\end{table*}

%% file: tabels/2_Ex1Result.tex
\begin{table*}[hbt]
    \centering
    \renewcommand{\arraystretch}{1.2}
    \caption{Performance comparison of different models based on Contact Detection and Touch Classification Metrics . ($RD = 0$ doesn't mean there is no recovery delay, but as the delay exceeds our threshold(150~ms), they will all be marked as FP)}
    \label{tab:exp1_results}

    \begin{tabular}{|c|c|ccccc|c|ccc|}
        \hline
        \multirow{3}{*}{\centering \textbf{Eperiment}} 
            & \multirow{3}{*}{\centering \textbf{Models}}
            & \multicolumn{5}{c|}{\textbf{Contact Detection Metrics}}
            & \multicolumn{4}{c|}{\textbf{Gesture Classification Metrics}} \\
        \cline{3-11}
        
        & & \multirow{2}{*}{\centering \textbf{Acc. (\%)}}
          & \multirow{2}{*}{\centering \textbf{DFR (\%)}}
          & \multirow{2}{*}{\centering \textbf{FAR (\%)}}
          & \multirow{2}{*}{\centering \textbf{DD (ms)}}
          & \multirow{2}{*}{\centering \textbf{RD (ms)}}
          & \multirow{2}{*}{\centering \textbf{Acc. (\%)}}
          & \multicolumn{3}{c|}{\centering \textbf{DFR (\%)}} \\  
        \cline{9-11}
        
        & & & & & & & & \textbf{ST} & \textbf{P} & \textbf{G} \\
        \hline
        
        \multirow{4}{*}{\centering EXP1}
        & LSTM(1 Layer) & 95.84 & 2.88 & 4.09 & 58.19 & 65 &   80.34 & 28.57 & 8.14 & 22.29 \\
        & STFT2DCNN & 95.98 & 3.01 & 3.65 & \textbf{45.12} & 75.00 &   \textbf{95.72} & \textbf{1.35}  & 9.12  & 2.37 \\
        & STFT3DCNN & 94.42 & 0.84 & 6.17 & 57.80 & 72.50 &   93.05 & 0     & 14.37 & 6.47 \\
        & STT2DCNN  & 96.42 & 0.47 & 4.39 & 64.64 & 0     &   88.76 & 25.00 & \textbf{6.49}  & \textbf{1.78} \\
        & STT3DCNN  & 94.69 & \textbf{0.13} & 6.22 & 61.83 & 75.00 &   94.26 & 3.23  & 7.49  & 6.50 \\
        & RT2DCNN   & \textbf{96.92} & 4.36 & \textbf{2.22} & 68.20 & \textbf{53.74} &   92.36 & 5.31  & 11.86 & 5.74 \\
        & RT3DCNN   & 96.12 & 0.81 & 4.30 & 61.00 & 81.66 &   91.58 & 9.52  & 8.77  & 6.95 \\
        \hline       
        \multirow{6}{*}{\centering EXP2}
        & LSTM(1 Layer) & 94.45 & 0.38 & 7.14 & 47.30 & 64.99& 74.90 & 38.57 & 8.85 & 27.87 \\
        & STFT2DCNN & 95.64 & 2.04 & 4.57 & \textbf{47.61} & 70.00 &   95.76 & 3.85  & 7.77  & \textbf{1.11} \\
        & STFT3DCNN & \textbf{96.22} & 1.64 & \textbf{4.15} & 51.45 & 56.67 &   93.08 & \textbf{1.27}  & 10.30 & 9.18 \\
        & STT2DCNN  & 95.81 & 1.26 & 4.43 & 52.12 & 39.99 &   92.25 & 8.16  & 10.14 & 4.95 \\
        & STT3DCNN  & 95.62 & 1.02 & 4.95 & 51.48 & 85.00 &   \textbf{95.92} & 2.78  & 7.81  & 1.65 \\
        & RT2DCNN   & 93.85 & 1.96 & 6.94 & 61.41 & 75.00 &   71.17 & 36.73 & 8.07  & 41.67 \\
        & RT3DCNN   & 92.59 & \textbf{0.19} & 8.39 & 51.89 & \textbf{32.51} &   71.67 & 45.83 & \textbf{4.23}  & 34.93 \\

        \hline
    \end{tabular}
\end{table*}

%% file: tabels/4_EXP3Comparison.tex
\begin{table}[t]
    \centering
    \renewcommand{\arraystretch}{1.2}
    \caption{Contact Detection (CD) and Gesture Classification (GC) performance comparison of 2D and 3D CNNs with different convolutional structures. In “Conv. Structures”, notation \(x+y\) denotes that the kernel size in the third dimension(the combined feature-joint indices dimension, as shown in Fig.~\ref{fig:input}) is set to \(x\) in the first convolutional layer and \(y\) in the second convolutional layer.}
    \label{tab:exp3_comparison}

    \begin{tabular}{|c|c|cc|}
        \hline
        \textbf{Models} & \textbf{Conv. Structures} & \textbf{CD Acc. (\%)} & \textbf{GC Acc. (\%)} \\
        
        \hline
        
        STFT2DCNN & 28+1 & 95.64 & 95.76 \\
        \hline
        \multirow{2}{*}{\centering STFT3DCNN}
        & 7+1 & 96.22 & 93.08 \\
        & 4+7 & - & - \\
        \hline
        STT2DCNN & 28+1 & 95.81 & 92.25 \\
        \hline
        \multirow{2}{*}{\centering STT3DCNN}
        & 7+1 & 92.50 & 86.05 \\
        & 4+7 & 95.62 & 95.92 \\
        \hline
    \end{tabular}
\end{table}